\documentclass[10pt,twocolumn,letterpaper]{article}

\usepackage{wacv}              % To produce the CAMERA-READY version
\usepackage{cuted}   % for strip
\usepackage{multirow}

\definecolor{wacvblue}{rgb}{0.21,0.49,0.74}
\usepackage[pagebackref,breaklinks,colorlinks,allcolors=wacvblue]{hyperref}

\def\wacvPaperID{3} % *** Enter the WACV Paper ID here
\def\confName{WACV}
\def\confYear{2027}

\title{File Size as a Potential Dataset Bias in Audio-Visual Deepfake Data}

\author{Marcella Astrid$^{\star,\dagger}$ \qquad Niki Maria Foteinopoulou$^{\star,\ddagger}$\\[0.5em]
$^{\star}$Interdisciplinary Centre for Security, Reliability and Trust (SnT),\\
University of Luxembourg, Luxembourg\\
$^{\dagger}$Helmholtz AI, Helmholtz Munich, Germany\\
$^{\ddagger}$Independent Researcher\\
Email: marcella.astrid@helmholtz-munich.de, nfoteinopoulou@gmail.com}

\begin{document}
\maketitle

\begin{strip}
\noindent\begin{minipage}{\textwidth}
\centering
\includegraphics[width=\linewidth]{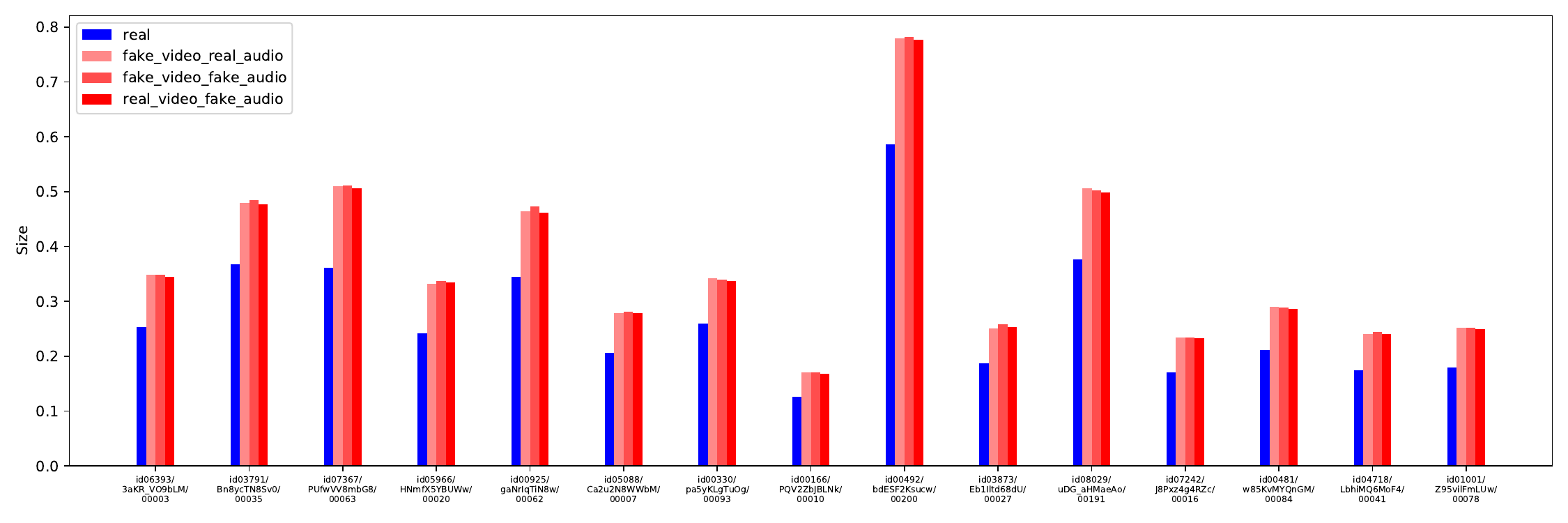}
\captionof{figure}{Samples of 15 real videos and their corresponding fake videos in the 1M-Deepfake Detection Challenge dataset. The x-axis represents the video name and the y-axis represents file size in MB. Blue and reddish colors represent the real and fake classes, respectively. In these examples, the fake outputs are generally larger than their corresponding real sources, illustrating a possible label-dependent file-size pattern.}
\label{fig:introduction_fig}
\end{minipage}
\end{strip}

\begin{abstract}
Deepfake detection benchmarks are commonly designed to evaluate whether models can distinguish authentic from manipulated audiovisual content. However, benchmark construction pipelines may introduce unintended associations between class labels and simple file-level properties. In this work, we identify a potential file-size bias in two widely used audio-visual deepfake datasets. In the 1M-Deepfakes Detection Challenge dataset, a score derived from file size alone achieves $70.13\%$ AUC on the validation set. When file size is considered together with video duration, simple file-level methods exceed $90\%$ AUC on 1M and $85\%$ AUC on FakeAVCeleb. Although the same pattern is not observed in DFDC, the results suggest that the observed association is dataset-dependent rather than an inherent property of deepfake videos. We discuss several possible explanations, including differences in encoding, compression, source data, manipulation procedures, and post-processing, although the present analysis does not establish the underlying cause. We further discuss whether the factors associated with file size may also leave traces in decoded video or audio, and consequently whether they could influence content-based deepfake detection, as an important direction for future work. Our findings highlight the importance of examining simple file-level properties when constructing and evaluating deepfake datasets and benchmarks.
\end{abstract}
    
\section{Introduction}
\label{sec:intro}

Recent generative techniques have greatly improved the realism of audio-visual deepfakes~\cite{prajwal2020lip,nirkin2019fsgan}. Although these techniques have many legitimate applications, manipulated media can also be used for fraud, impersonation, and misinformation. For example, a finance worker was reportedly deceived into transferring 25 million US dollars after participating in a video call involving deepfake impersonation~\cite{Scam_news}. Reliable deepfake detection and trustworthy evaluation benchmarks are therefore increasingly important.

Most work in audio-visual deepfake detection focuses on designing stronger models that exploit visual, audio, or cross-modal evidence~\cite{yang2023avoid,wang2024avt2,zhou2021joint,salvi2023robust}. Such progress implicitly assumes that benchmark performance primarily reflects a model's ability to recognize manipulation-related evidence. However, this assumption can be violated when the dataset construction process introduces auxiliary cues that correlate with the class label. A detector can then achieve high benchmark performance by exploiting a shortcut that is specific to the dataset pipeline rather than to deepfake content itself.

In this work, we investigate a potential file-size bias in audio-visual deepfake datasets by examining the relationship between file size, video duration, and the real/fake label. We observe a strong association in the 1M-Deepfakes Detection Challenge dataset. Figure~\ref{fig:introduction_fig} provides an initial illustration of this pattern using paired real and fake samples, where the selected fake outputs are generally larger than their corresponding real sources. We further show that considering file size relative to video duration reveals an even stronger association with the class label.

File size can reflect several aspects of video construction and processing, including duration, encoding, bitrate, audio coding, and post-processing. The strong association observed in 1M therefore motivates a broader investigation of whether similar file-level patterns occur across deepfake datasets. Whether the underlying differences are also reflected in decoded audiovisual content remains an important question for future work.

Our contributions are as follows:
\begin{itemize}
    \item We identify and quantify a strong association between file size, video duration, and the real/fake label in 1M. File size alone achieves $70.13\%$ AUC, while considering file size relative to video duration increases the AUC to over $91\%$.

    \item We compare 1M, FakeAVCeleb, and DFDC and find that the observed file-size pattern is present in 1M and FakeAVCeleb but not evident in DFDC.

    \item We perform cross-dataset evaluation between 1M and FakeAVCeleb and find that similar relationships between file-level properties and class labels are observed across the two datasets.

    \item We discuss possible explanations for the observed patterns, including differences in encoding, compression, source data, and post-processing, and identify as future work whether the underlying differences are also reflected in decoded audiovisual
    content.
\end{itemize}

The remainder of the paper is organized as follows. Section~\ref{sec:background} provides background on deepfake benchmark evaluation, discusses the relationship between file size and video duration, and formulates the research questions investigated in this work. Section~\ref{sec:preliminary_findings} presents the initial evidence of file-size bias. Section~\ref{sec:methodology} describes how we quantify the relationship between file size, video duration, and the class labels. Section~\ref{sec:experiments} describes the datasets and experimental setup, evaluates the bias within 1M, examines its presence in FakeAVCeleb and DFDC, and analyzes its cross-dataset behavior. Finally, we discuss the interpretation and limitations of the findings, their implications for deepfake benchmark design, and directions for future work before concluding the paper.
\section{Background and motivation}
\label{sec:background}

\subsection{Deepfake detection and benchmark evaluation}
Audio-visual deepfake detectors increasingly combine information from facial appearance, temporal dynamics, speech, and cross-modal correspondence~\cite{yang2023avoid,wang2024avt2,zhou2021joint,salvi2023robust}. These approaches are typically compared on benchmark datasets that provide a binary real/fake label. Consequently, the interpretation of detector performance depends not only on the model but also on how the benchmark itself was assembled. Ideally, the label should primarily reflect whether the semantic or forensic content has been manipulated. In practice, however, real and fake samples may pass through different generation, rendering, compression, or post-processing pipelines.

This distinction is particularly relevant for synthetic-media datasets because fake samples are often produced from real source videos and then written to disk again. Re-saving a manipulated sample can change compression statistics even when the manipulation occupies only a small spatial or temporal region. If such processing differs systematically between classes, variables unrelated to the manipulation target can become predictive of the label. A benchmark can therefore contain unintended label-associated information that was not intended as part of the detection task.

\subsection{File size and video duration}

File size is a simple file-level property that can be examined without accessing the audiovisual content itself. It is influenced by several characteristics of a video, including its duration, bitrate, codec configuration, audio coding, frame complexity, and container overhead. Consequently, differences in the construction or processing of real and fake samples may also be reflected in their file sizes.

A systematic association between file size and the real/fake label can therefore reveal differences between the two class distributions that are not apparent from the labels alone. Examining this association provides a simple way to characterize the dataset and to investigate whether file-level properties differ consistently between authentic and manipulated samples. Since video duration naturally affects file size, we additionally consider the relationship between these two variables rather than examining absolute file size alone.

\subsection{Research questions}
We organize the study around four research questions:
\begin{itemize}
    \item \textbf{RQ1:} To what extent is video file size associated with the real/fake label in the 1M dataset?
    \item \textbf{RQ2:} Does considering file size relative to video duration reveal a stronger association with the real/fake label in 1M than file size alone?
    \item \textbf{RQ3:} Is the file-size pattern observed in 1M also present in other audio-visual deepfake datasets?
    \item \textbf{RQ4:} Do the relationships between file size, video duration, and class labels transfer between datasets?
\end{itemize}

These questions can be answered with the measurements already available in the datasets and do not require assumptions about the internal behavior of a deepfake detector. Whether the same underlying bias is also exploitable from decoded audiovisual content is a separate question, which we discuss as future work in Section~\ref{subsec:discussion}.

\section{Initial evidence of file-size bias}
\label{sec:preliminary_findings}

Figure~\ref{fig:introduction_fig} provides an initial illustration of file-size differences between corresponding real and fake videos in 1M. In the selected examples, the manipulated outputs generally occupy more storage than their corresponding real sources. While these examples motivate our investigation, they alone do not establish whether the pattern is present across the dataset.

\begin{figure}[tb]
\centering
\begin{subfigure}{0.49\linewidth}
    \includegraphics[width=\linewidth]{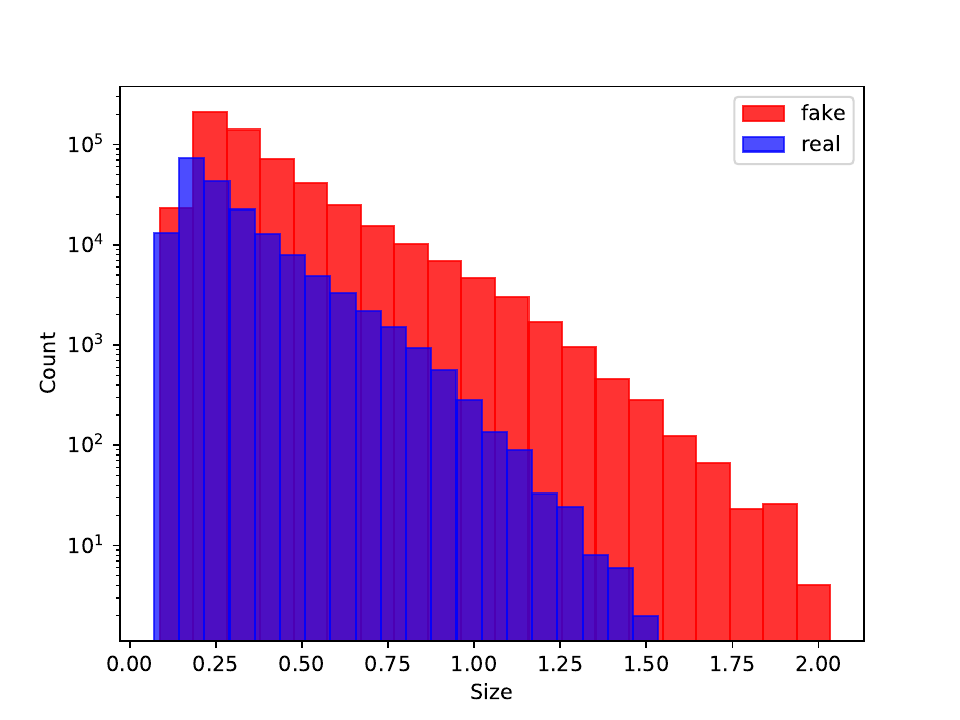}
    \caption{}
    \label{fig:train_storage}
\end{subfigure}
\hfill
\begin{subfigure}{0.49\linewidth}
    \includegraphics[width=\linewidth]{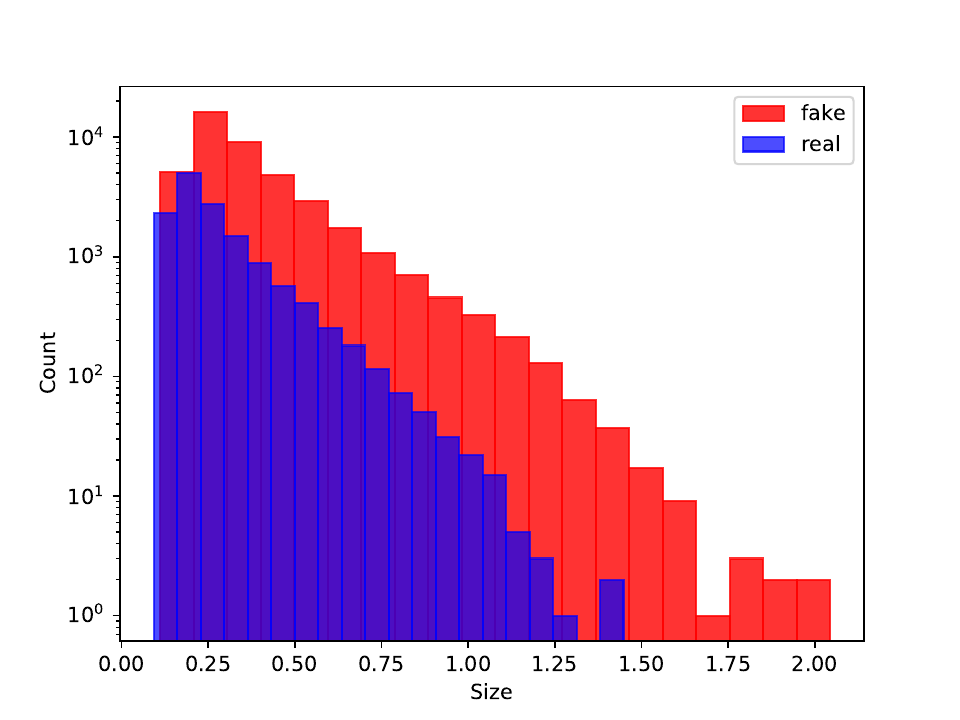}
    \caption{}
    \label{fig:val_storage}
\end{subfigure}
\caption{Real and fake file-size distributions for (a) the training set and (b) the validation set of 1M. The x-axis shows video size (MB) and the y-axis shows the number of videos. Blue and red indicate real and fake classes, respectively. The separation indicates label-dependent differences in the dataset's file-level statistics.}
\label{fig:storage_distribution}
\end{figure}

We therefore examine the class-conditional file-size distributions in
Figure~\ref{fig:storage_distribution}. The real and fake distributions are partially separable in both the training and validation splits. This does not imply that every fake video is larger than every real video, nor would such a deterministic separation be expected. Although file size does not perfectly separate real and fake videos, the differences between their file-size distributions are sufficiently strong to carry information about the class label.

We first measure how predictive file size is without using any learned audiovisual representation. Let $s$ denote file size. We assign larger files a higher fake score by min-max normalizing the validation file size using statistics from the training set:
\begin{equation}
    y' = \frac{s^{val} - s^{train}_{min}}{s^{train}_{max} - s^{train}_{min}} \text{,}
\end{equation}
where $s^{val}$ is the file size of a validation video and $s^{train}_{min}$ and $s^{train}_{max}$ are the minimum and maximum file sizes in the training split. File size is measured in megabytes (MB). We clip the score to $[0,1]$:
\begin{equation}
    y = \min(1, \max(0, y')) \text{.}
\end{equation}

This trivial score reaches $70.13\%$ AUC on the 1M validation set. Because the score uses no pixels, audio samples, facial information, synchronization cues, or learned deepfake representation, the result shows that file size alone is associated with the benchmark label. The AUC formulation is useful here because it measures ranking ability without requiring us to select an arbitrary file-size threshold. In other words, larger files are sufficiently more common among fake samples that file size alone ranks real and fake videos substantially better than chance.

This result answers RQ1 for 1M, showing that file size alone carries information about the real/fake label. We next examine whether accounting for video duration strengthens this association.

\begin{figure}[bpt]
\centering
\begin{subfigure}{0.32\linewidth}
    \includegraphics[width=\linewidth]{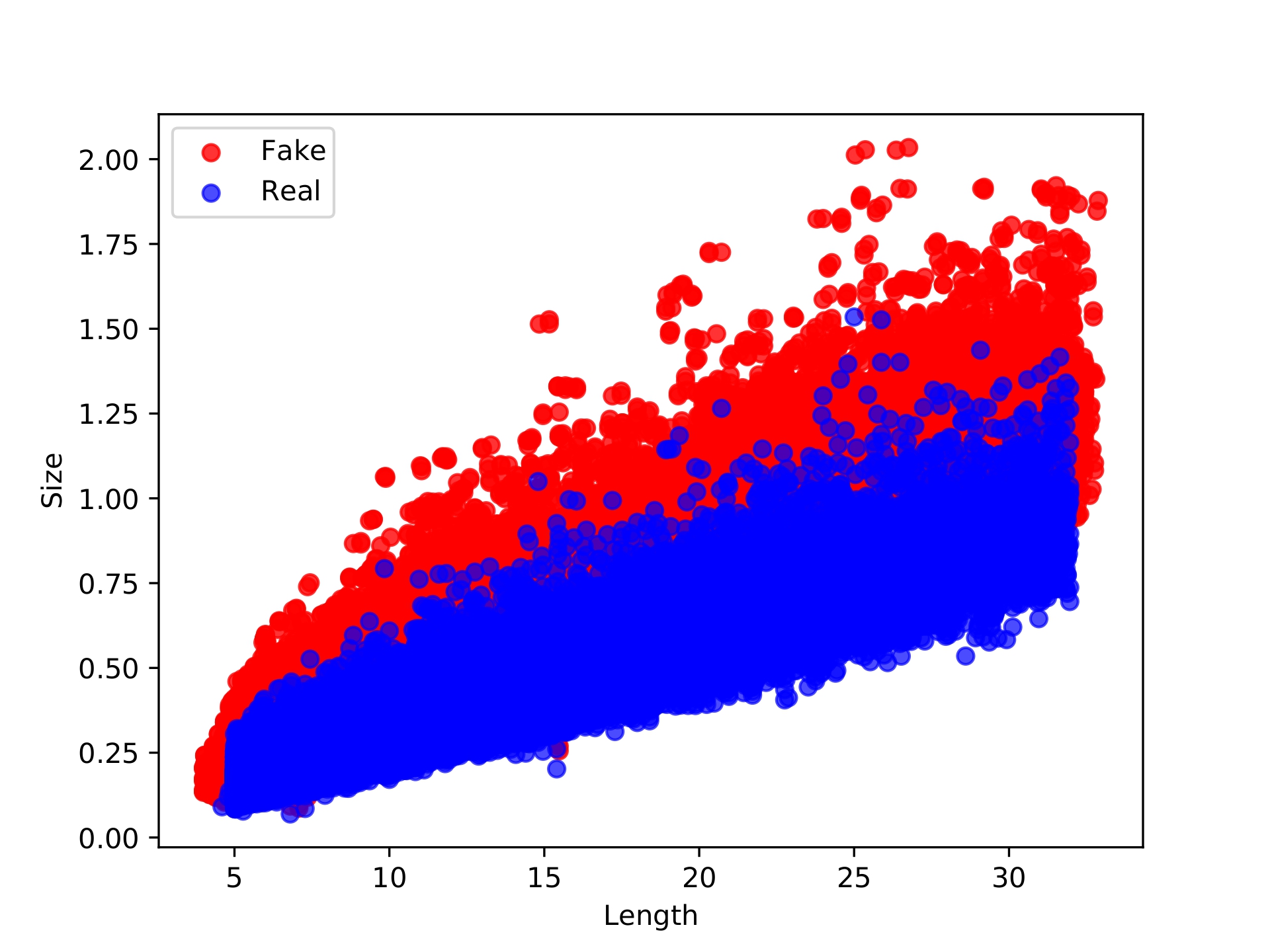}
    \caption{}
    \label{fig:train_mb_len}
\end{subfigure}
\hfill
\begin{subfigure}{0.32\linewidth}
    \includegraphics[width=\linewidth]{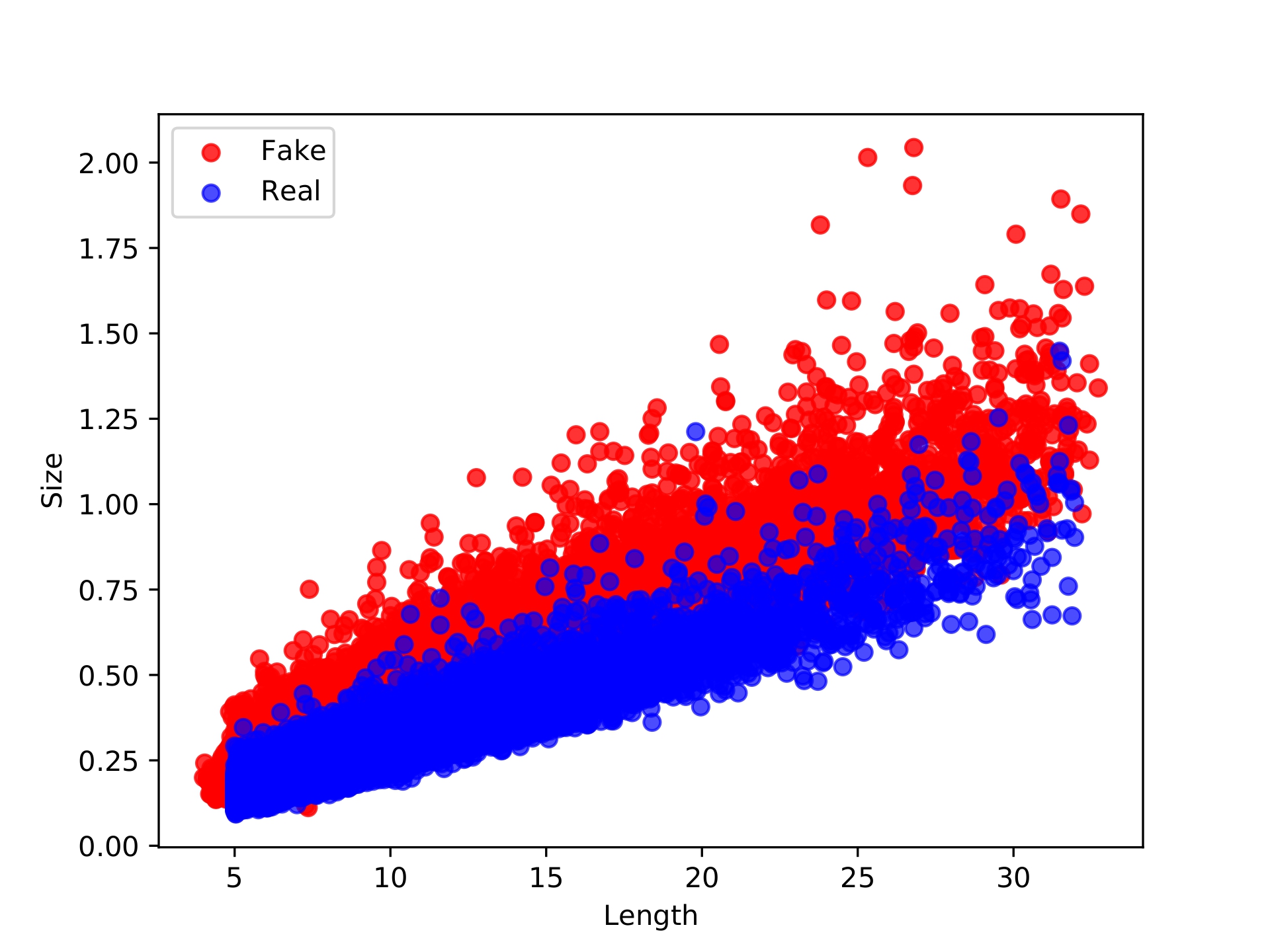}
    \caption{}
    \label{fig:val_mb_len}
\end{subfigure}
\hfill
\begin{subfigure}{0.32\linewidth}
    \includegraphics[width=\linewidth]{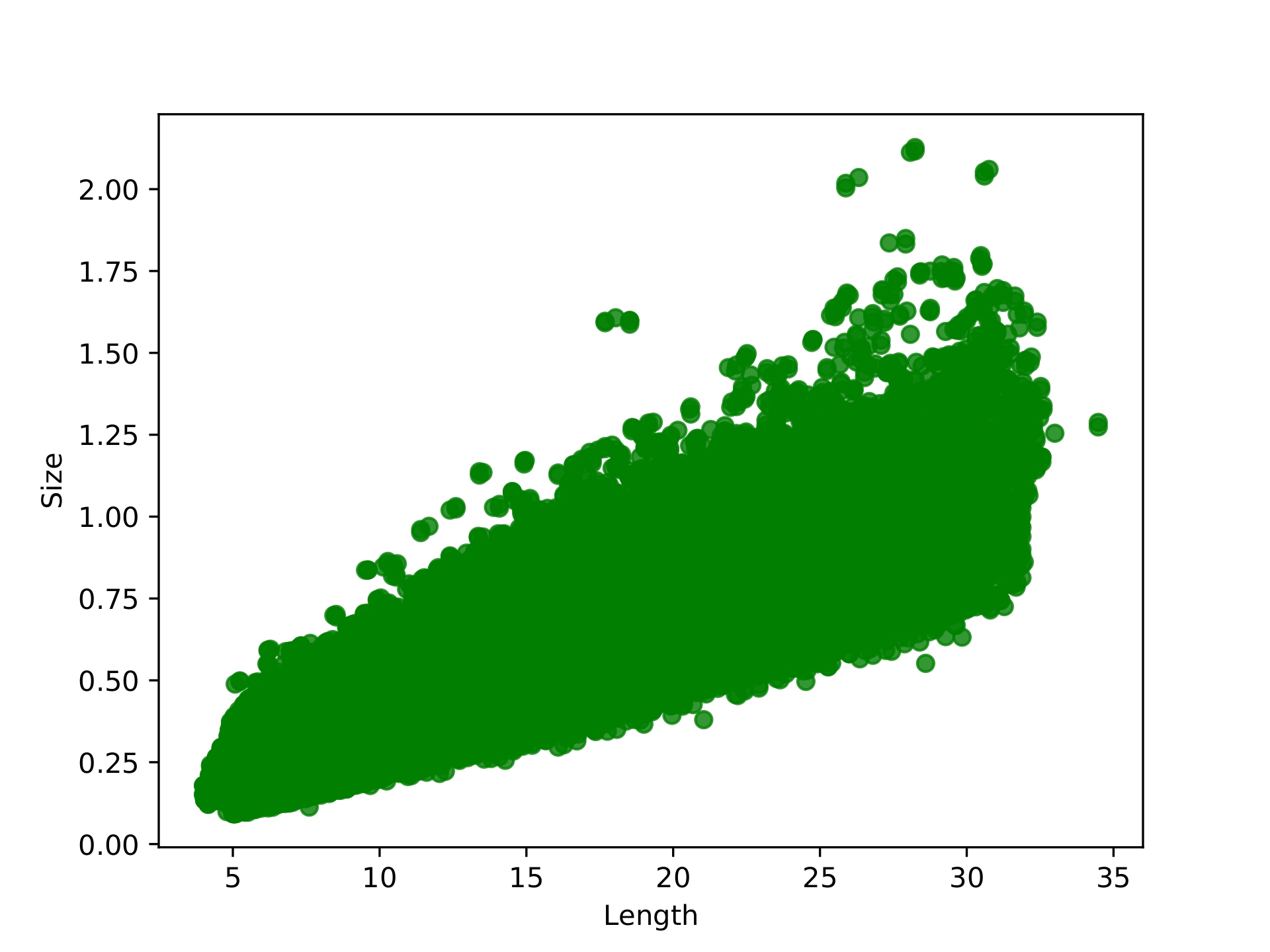}
    \caption{}
    \label{fig:test_mb_len}
\end{subfigure}
\caption{File size (y-axis) versus video duration (x-axis) for (a) the training set, (b) the validation set, and (c) the test set of 1M. Blue points denote real videos, red points denote fake videos, and green points denote samples with unknown labels. In the labeled training and validation sets, fake videos tend to have larger file sizes than real videos of similar duration.}
\label{fig:mb_len_distribution}
\end{figure}

\section{Quantifying file-size bias}
\label{sec:methodology}

The previous section shows that file size alone is associated with the real/fake label in 1M. However, file size naturally depends on video duration: longer videos generally require more storage. We therefore consider file size together with video duration to better characterize the observed association. Figure~\ref{fig:mb_len_distribution} shows that the relationship between these two variables differs between the real and fake classes.

We quantify this relationship in two ways. First, we consider file size relative to video duration through a simple file-size-per-second measure. Second, we use logistic regression and support vector classifiers to model file size and duration jointly. The resulting AUC values provide a quantitative measure of how strongly these simple file-level properties are associated with the class labels.

\begin{figure}[tb]
\centering
\begin{subfigure}{0.49\linewidth}
    \includegraphics[width=\linewidth]{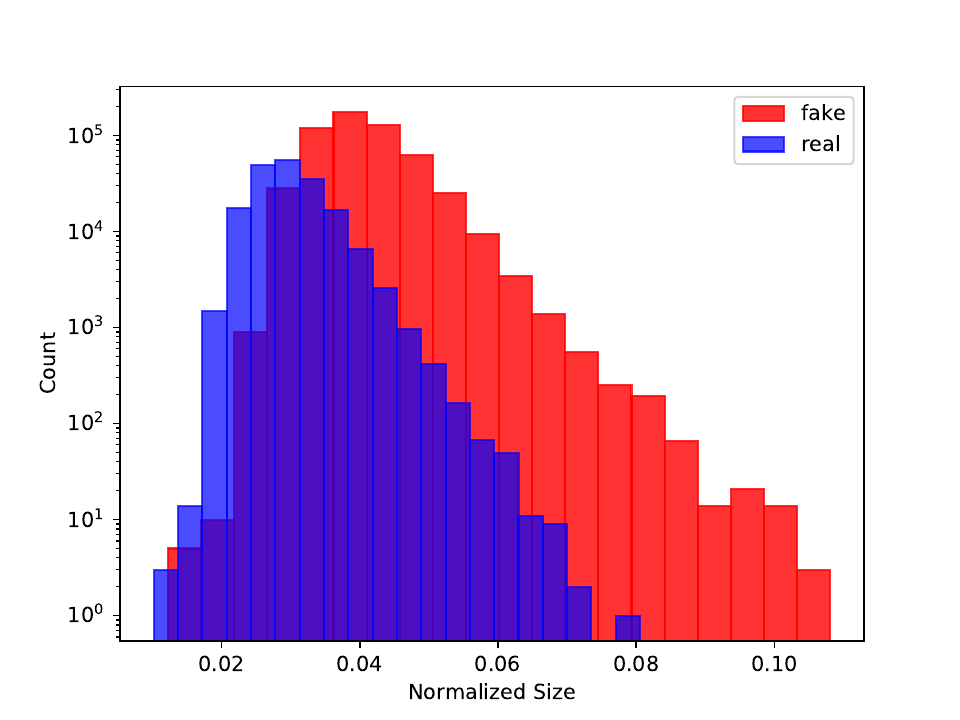}
    \caption{}
    \label{fig:train_storage_lennormalized}
\end{subfigure}
\hfill
\begin{subfigure}{0.49\linewidth}
    \includegraphics[width=\linewidth]{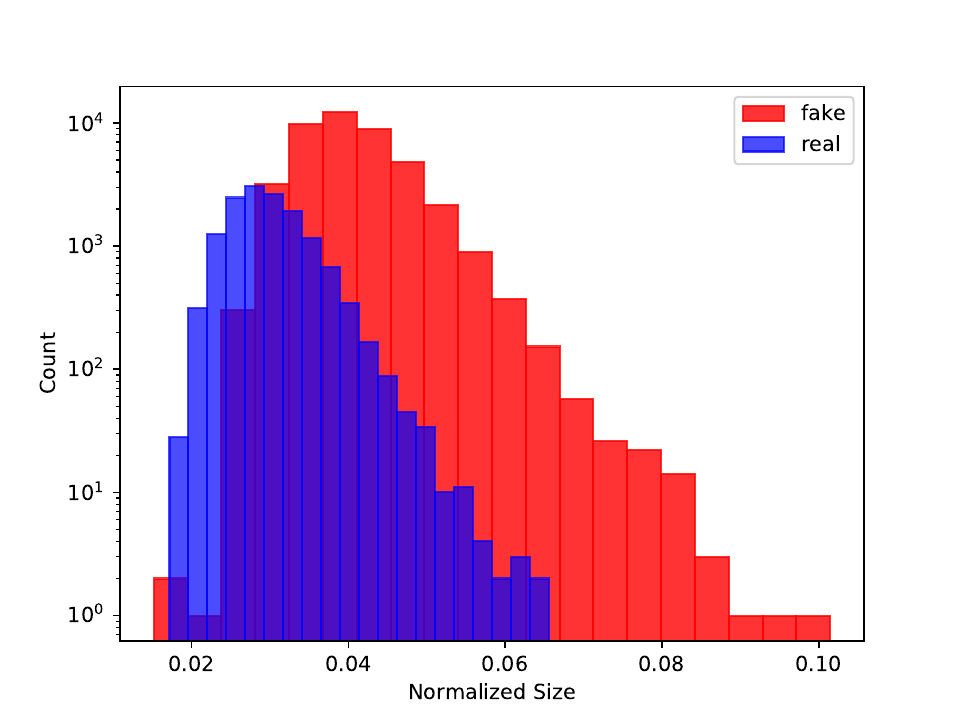}
    \caption{}
    \label{fig:val_storage_lennormalized}
\end{subfigure}
\caption{Real and fake distributions of file size per second $n$ from Eq.~\eqref{eq:size_per_second} on (a) the training set and (b) the validation set of 1M. The x-axis label ``Normalized Size'' refers to the ratio between file size and video duration, measured in MB/s. Blue and red denote the real and fake classes, respectively. The real and fake distributions are more clearly separated than the file-size-only distributions in Fig.~\ref{fig:storage_distribution}.}
\label{fig:storage_distribution_lennormalized}
\end{figure}

\subsection{File size and video duration}
\label{subsec:input}

Let $s$ denote the file size of a video in megabytes (MB) and $l$ its duration in seconds. We represent each video using the two-dimensional vector
\begin{equation}
    \mathbf{x} =
    \begin{bmatrix}
        s \\
        l
    \end{bmatrix}.
\end{equation}

Figure~\ref{fig:mb_len_distribution} plots file size against video duration for the 1M training, validation, and test splits. As expected, file size generally increases with video duration. More importantly, the labeled training and validation data show different size--duration relationships for real and fake videos. The class separation therefore remains visible when file size is considered jointly with duration, motivating a more direct analysis of the relationship between the two variables.

\subsection{File size per second}
\label{subsec:size_per_second}

A simple way to account for video duration is to consider the amount of storage occupied per second of video. We define
\begin{equation}
    n = \frac{s}{l},
    \label{eq:size_per_second}
\end{equation}
where $s$ is measured in MB and $l$ in seconds. Thus, $n$ is measured
in MB/s.

To obtain a score in the range $[0,1]$, we min--max normalize $n$ using the minimum and maximum values observed in the training set:
\begin{equation}
    y' =
    \frac{n-n^{train}_{min}}
         {n^{train}_{max}-n^{train}_{min}},
\end{equation}
followed by
\begin{equation}
    y = \min(1,\max(0,y')).
\end{equation}
Here, $n^{train}_{min}$ and $n^{train}_{max}$ denote the minimum and
maximum file-size-per-second values in the training data,
respectively.

Figure~\ref{fig:storage_distribution_lennormalized} shows the class-conditional distributions of file size per second for the 1M training and validation sets. Compared with file size alone (Fig.~\ref{fig:storage_distribution}), the distributions exhibit stronger separation between the real and fake classes. This suggests that the relationship between file size and video duration carries considerably more information about the class label than absolute file size alone.

\subsection{Logistic regression and support vector classifiers}
\label{subsec:classifiers}

File size per second imposes a fixed relationship between file size and duration. To complement this measure, we also consider classifiers that learn the relationship between the two variables directly from the training data.

We first use logistic regression (LR), which models the real/fake score as
\begin{equation}
    y =
    \frac{1}
    {1+\exp^{-(\mathbf{w}^{T}\mathbf{x}+b)}},
    \label{eq:sigmoid}
\end{equation}
where $\mathbf{x}=[s,l]^{T}$ and the parameters $\mathbf{w}$ and $b$
are learned from the training data.

We additionally use a support vector classifier (SVC) to model the separation between the two classes in the same size--duration space. Since the FakeAVCeleb training set is highly imbalanced, we also evaluate a balanced SVC (B-SVC), which assigns class weights according to the training-set class frequencies.

These approaches characterize the file-size bias from complementary perspectives. File size per second directly measures storage relative to video duration, whereas LR and SVC treat file size and duration as separate variables and learn their relationship with the class label from the training data. In the following section, we compare their AUC values across datasets to assess the strength and consistency of the observed file-size bias.

\section{Experiments}
\label{sec:experiments}

\subsection{Datasets}
\subsubsection{1M-Deepfake Detection Challenge Dataset (1M)}
The 1M dataset~\cite{cai2023av} contains more than one million videos from over 2,000 subjects, with audio-only, video-only, and audio-visual manipulations. Manipulations are applied to portions of the video rather than necessarily to the entire clip. We use the official train, validation, and test splits and refer to the dataset as \textit{1M}.

\subsubsection{FakeAVCeleb (FAV)}
FakeAVCeleb~\cite{khalid2021fakeavceleb} contains audio-only, video-only, and audio-visual manipulations. Following~\cite{khalid2021evaluation}, we construct a balanced test set of 70 real and 70 fake videos. The training set excludes identities present in the test set and contains 430 real and 18,018 fake videos. We refer to this dataset as \textit{FAV}.

\subsubsection{DeepFake Detection Challenge (DFDC)}
DFDC~\cite{dolhansky2020deepfake} contains more than 100,000 clips and 3,426 identities and is primarily based on face swaps. Following~\cite{chugh2020not, mittal2020emotions}, we use a subset of 18,000 videos and split it into training and test sets with an 85:15 ratio.

\begin{table}[]
\caption{AUC (\%) obtained using only file size and video duration. Results are reported on 1M validation (1M-val), 1M test (1M-test), and FAV test (FAV-test), using 1M or FAV as the training/reference dataset. High AUC indicates a strong association between these file-level properties and the class labels. Bold and underline denote the best and second-best values in each column.}
\resizebox{\linewidth}{!}{
\begin{tabular}{|l|ccc|cc|}
\hline
\multirow{3}{*}{Method}   & \multicolumn{3}{c|}{1M-train}                    & \multicolumn{2}{c|}{FAV-train}                   \\ \cline{2-6}
                          & 1M-val & 1M-test & FAV-test & 1M-val & FAV-test \\
                          & (a) & (b) & (c) & (d) & (e) \\\hline
Size/Second & \textbf{91.20} & \textbf{91.47} & \textbf{85.73} & \textbf{91.20} & \textbf{85.76} \\
LR       & 90.73 & 90.87 & 84.86 & \underline{86.52} & \underline{85.61} \\
SVC      & 91.17 & 91.36 & 85.16 & 47.16 & 41.20 \\
B-SVC    & \underline{91.19} & \underline{91.38} & \underline{85.20} & 85.91 & 84.78 \\ \hline
\end{tabular}
\label{tab:results}
}
\end{table}

\subsection{Experimental Setup}
Our code is written in Python 3.7 using scikit-learn 1.0.2 for SVC\footnote{\url{https://scikit-learn.org/1.0/modules/generated/sklearn.svm.SVC.html}} and logistic regression\footnote{\url{https://scikit-learn.org/1.0/modules/generated/sklearn.linear_model.LogisticRegression.html}}. Default parameters are used except for B-SVC, where \textit{class\_weight} is set to \texttt{balanced}. We report video-level Area Under the ROC Curve (AUC). We use only file size and video duration to quantify their association with the class labels. For Size/Second, the min--max normalization parameters are obtained from the corresponding training/reference dataset.

\subsection{Strength of the bias in 1M}
Table~\ref{tab:results}(a--b) shows the results on the 1M validation and test splits. All approaches achieve approximately $90\%$ AUC or higher, with Size/Second reaching $91.20\%$ on validation and $91.47\%$ on test. Compared with the $70.13\%$ AUC obtained using file size alone in Section~\ref{sec:preliminary_findings}, considering file size relative to video duration substantially strengthens the association with the real/fake label. This answers RQ2 and shows that considering file size relative to video duration provides substantially stronger class separation than absolute file size alone. The similar validation and test results further indicate that this pattern is consistent across the two 1M splits.

% Figure~\ref{fig:classifier} visualizes the classification boundaries learned by LR, SVC, and B-SVC. Their simple boundaries are sufficient to separate a substantial fraction of real and fake samples, reinforcing the interpretation that 1M contains a systematic label-dependent storage pattern.

% \begin{figure}[tb]
% \centering
% \begin{subfigure}{0.49\linewidth}
%     \includegraphics[width=\linewidth]{figures/train_logisticreg_classifier.jpg}
%     \caption{}
%     \label{fig:train_logisticreg_classifier}
% \end{subfigure}
% \hfill
% \begin{subfigure}{0.49\linewidth}
%     \includegraphics[width=\linewidth]{figures/train_svc_classifier.jpg}
%     \caption{}
%     \label{fig:train_svc_classifier}
% \end{subfigure}
% \hfill
% \begin{subfigure}{0.49\linewidth}
%     \includegraphics[width=\linewidth]{figures/train_svcbalanced_classifier.jpg}
%     \caption{}
%     \label{fig:train_svcbalanced_classifier}
% \end{subfigure}
% \caption{Classification boundaries of diagnostic probes on the 1M training data: (a) logistic regression, (b) support vector classifier, and (c) balanced support vector classifier. The purpose of these classifiers is to visualize and quantify the size--duration bias rather than to serve as deployable deepfake detectors.}
% \label{fig:classifier}
% \end{figure}

\subsection{Analysis on FAV and DFDC}
\begin{figure}[tb]
\centering
\begin{subfigure}{0.49\linewidth}
    \includegraphics[width=\linewidth]{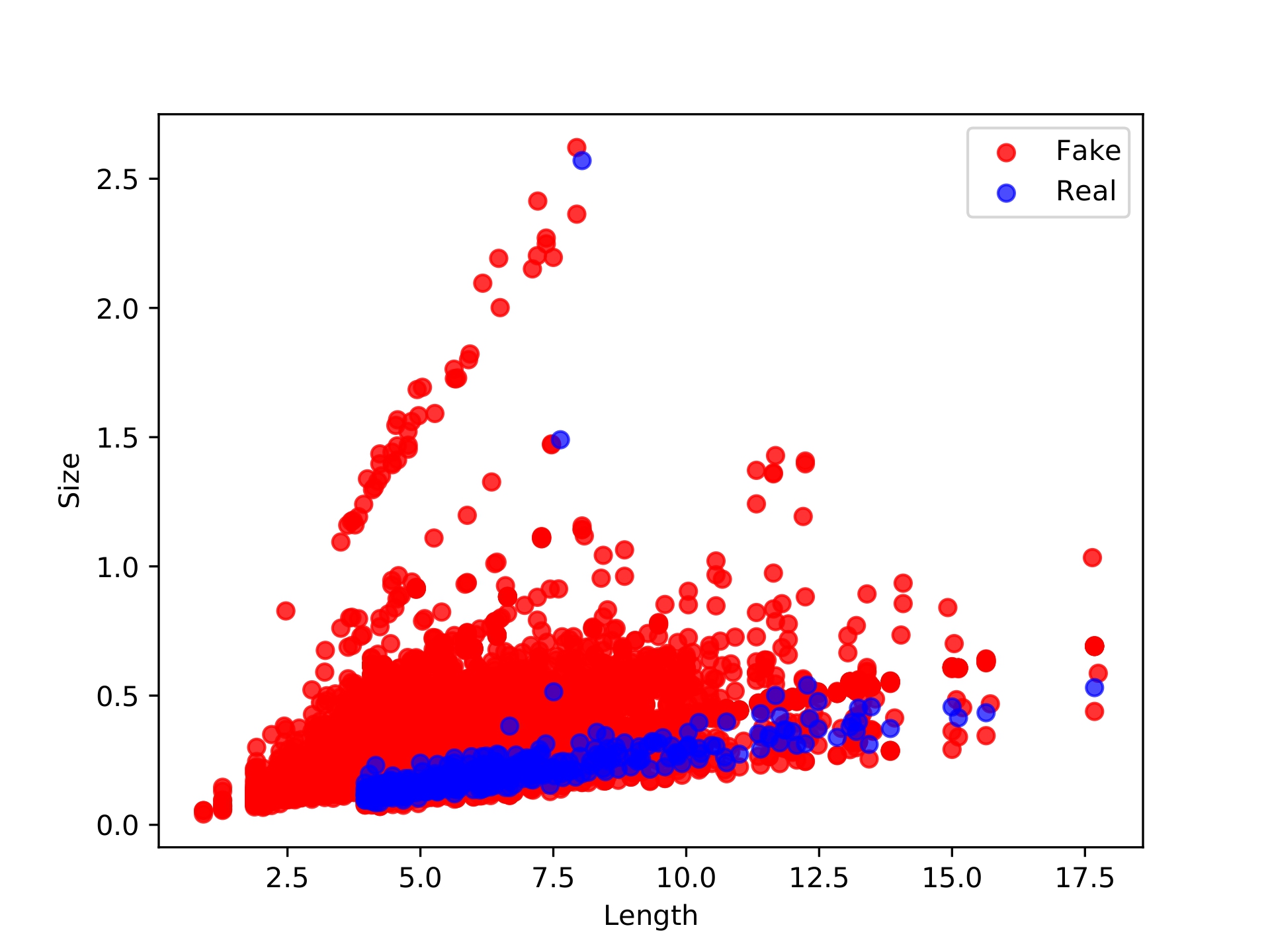}
    \caption{}
    \label{fig:fav_train_mb_len}
\end{subfigure}
\hfill
\begin{subfigure}{0.49\linewidth}
    \includegraphics[width=\linewidth]{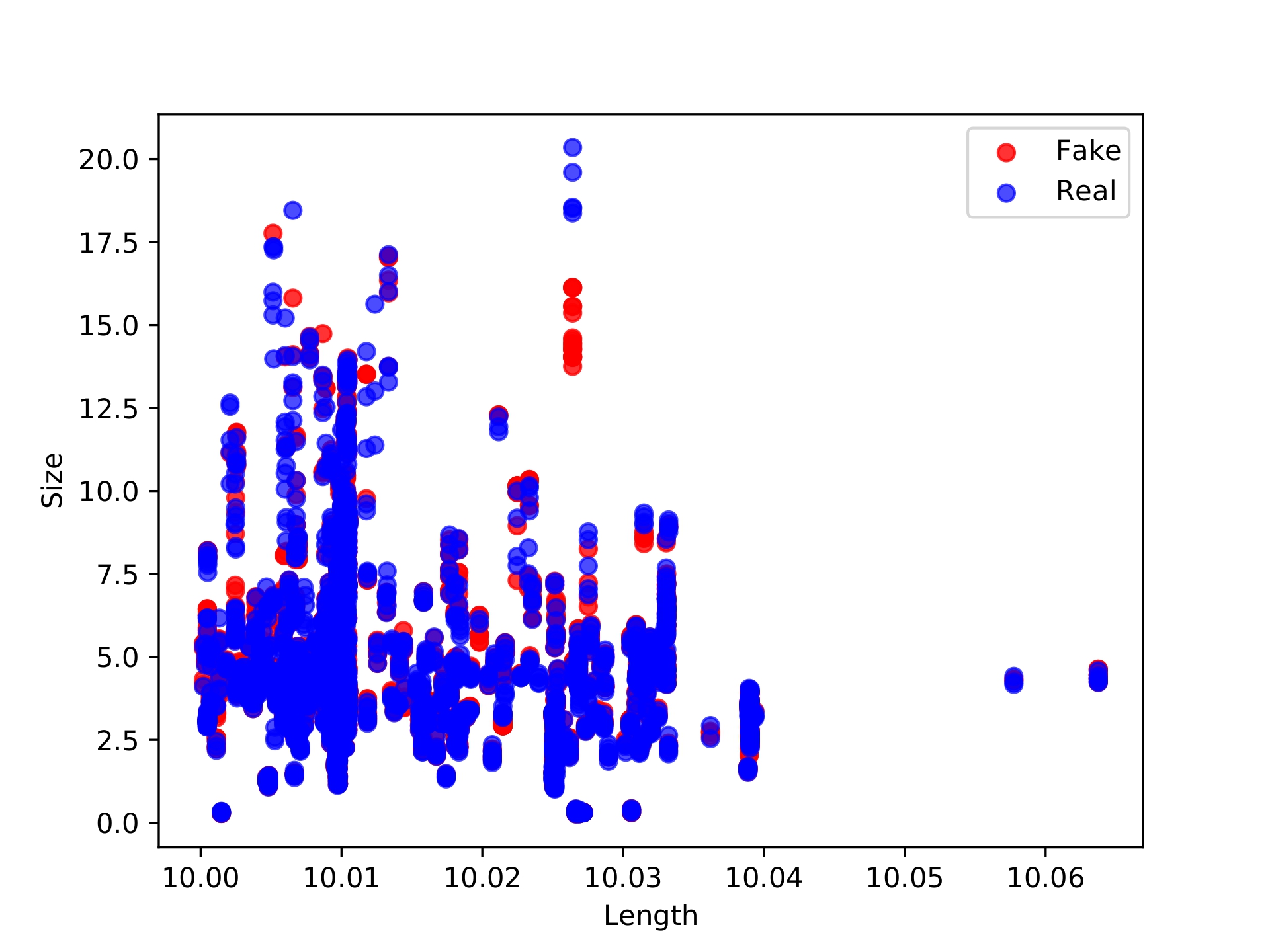}
    \caption{}
    \label{fig:dfdc_train_mb_len}
\end{subfigure}
\caption{File size (y-axis) versus video duration (x-axis) for the training sets of (a) FAV and (b) DFDC. FAV exhibits a class-dependent structure similar to 1M, whereas DFDC does not show an obvious separation. This contrast suggests that the observed file-size pattern is dataset-dependent rather than a general characteristic of deepfake media.}
\label{fig:other_mb_len_distribution}
\end{figure}

We next test whether the same file-size pattern appears in other audio-visual deepfake datasets. Applying the file-size-only score from Section~\ref{sec:preliminary_findings} yields $66.34\%$ AUC on FAV~\cite{khalid2021fakeavceleb} but only $49.97\%$ on DFDC~\cite{dolhansky2020deepfake}. Figure~\ref{fig:other_mb_len_distribution} similarly shows a class-dependent size--duration structure in FAV but no clear separation in DFDC.

This contrast suggests that the observed file-size association is dataset-dependent rather than an inherent property of deepfake media. Table~\ref{tab:results}(e) further shows that, within FAV, size and duration are sufficient to obtain approximately $85\%$ AUC for Size/Second, LR, and B-SVC. The unbalanced SVC performs poorly, plausibly because the FAV training split is highly imbalanced.

DFDC provides an informative contrasting case. Its file-size-only result is essentially random, and Figure~\ref{fig:other_mb_len_distribution}(b) does not exhibit the clear class-dependent pattern observed in 1M and FAV. This indicates that the observed association is present in some datasets but is not universal, answering RQ3.

\subsection{Cross-dataset analysis}
\label{subsec:cross_dataset}

We further examine whether the relationship between file size, video duration, and the class labels is similar across datasets. Table~\ref{tab:results}(c) reports results when the models are trained on 1M and evaluated on FAV, while Table~\ref{tab:results}(d) reports results when the models are trained on FAV and evaluated on 1M.

When trained on 1M and evaluated on FAV, LR, SVC, and B-SVC achieve AUCs between $84.86\%$ and $85.20\%$. These results are comparable to their performance when trained and evaluated on FAV, except for SVC, which performs poorly when trained on the highly imbalanced FAV training set. In the reverse direction, LR and B-SVC trained on FAV achieve $86.52\%$ and $85.91\%$ AUC, respectively, when evaluated on 1M.

Overall, the cross-dataset results show that several relationships learned from file size and video duration transfer between 1M and FAV, answering RQ4. The underlying reason for this similarity remains unclear and is discussed further in Section~\ref{subsec:discussion}.

\subsection{Discussion, limitations, and future work}
\label{subsec:discussion}

\subsubsection{What the current results establish}
\label{subsubsec:current_results}

Our results show that file-level properties are strongly associated with the real/fake labels in 1M and FAV. In 1M, file size alone provides $70.13\%$ AUC, while considering file size relative to video duration increases the AUC to over $91\%$. A similar pattern is observed in FAV, where file size alone achieves $66.34\%$ AUC and methods using file size and duration achieve approximately $85\%$ AUC. In contrast, the file-size-only result on DFDC is close to random, and its size--duration distribution does not show the clear class-dependent structure observed in 1M and FAV. Together, these results suggest a dataset-dependent file-size bias rather than a general characteristic of manipulated media.

The exact cause of this pattern remains unclear. One possible explanation is that real and fake videos undergo different processing or encoding steps during dataset construction. In both 1M and FAV, manipulated samples are generated from existing source videos~\cite{cai2023av,khalid2021fakeavceleb}. The generation process may introduce additional steps that contribute to differences in file size, although our experiments do not establish the underlying cause. Other factors, including codec settings, compression parameters, audio encoding, frame complexity, and containerization, may also contribute.

Previous video-forensics studies have shown that compression history can leave detectable traces in encoded videos. For example, double compression has been identified from quantization errors and coding-related statistics~\cite{aghamaleki2016inter,uddin2022double}. These findings do not explain the file-size patterns observed in our datasets, but they show that differences in processing history can leave measurable characteristics beyond file size itself.

\subsubsection{Future work}
Whether the observed file-size patterns are also reflected in decoded audiovisual content remains an open question. Differences in rate control, quantization, codec settings, audio encoding, or repeated compression can affect both storage size and the decoded signal. Future work can investigate this relationship through controlled re-encoding, file-level-matched evaluation subsets, and before/after evaluation of content-based detectors. Such analyses would help determine whether the observed file-size patterns reflect broader processing differences that are also accessible from audiovisual content.

\subsubsection{Implications for dataset design}
The results motivate examining simple file-level properties when releasing synthetic-media benchmarks. At minimum, dataset creators can compare class distributions of duration, file size, bitrate, resolution, frame rate, codec, and audio encoding. These checks may be particularly useful when authentic and manipulated samples undergo different collection or processing procedures.

When strong class associations are found, benchmark providers can investigate and, where appropriate, control the relevant file-level property, construct matched evaluation splits, or explicitly document the observed association. Researchers using existing benchmarks can likewise examine whether evaluation results remain stable across file-level-matched subsets. Such practices can make dataset-level differences more visible and reported performance easier to interpret.

\section{Conclusion}
\label{sec:conclusion}

This paper investigates the relationship between file-level properties and class labels in audio-visual deepfake datasets. We find that file size alone carries information about the real/fake label in 1M and FakeAVCeleb, and that considering file size together with video duration reveals a substantially stronger association. Simple methods using these properties achieve over $90\%$ AUC on 1M and approximately $85\%$ AUC on FakeAVCeleb. In contrast, the same file-size pattern is not observed in DFDC, suggesting that the observed association is dataset-dependent rather than an intrinsic characteristic of deepfake media.

The underlying cause of these patterns remains unclear. Differences in source data, generation, encoding, compression, and post-processing are possible contributors, but the present analysis does not determine which factors are responsible. Future work should investigate whether the differences associated with file size are also reflected in the decoded audiovisual content and whether they influence content-based deepfake detection. More broadly, our findings highlight the importance of examining simple file-level properties when constructing and evaluating deepfake benchmarks, as these properties may reveal unexpected associations with the class labels.

\section*{Acknowledgment}
This work was supported by the Luxembourg National Research Fund (FNR) under the project BRIDGES2021/IS/16353350/FaKeDeTeR and POST Luxembourg. The experiments were conducted while Marcella Astrid and Niki Maria Foteinopoulou were at the University of Luxembourg, and the manuscript was prepared while Marcella Astrid was with Helmholtz AI, Helmholtz Munich. We thank Marie Piraud (Team Lead, AI Consultants for Health, Helmholtz AI) for her support in submitting this work.

{
    \small
    \bibliographystyle{ieeenat_fullname}
    \bibliography{main}
}

\end{document}